\documentclass[final,onefignum,onetabnum]{siamonline250211}
\usepackage{lipsum}
\usepackage{amsfonts}
\usepackage{graphicx}
\usepackage{epstopdf}
\usepackage{algorithmic}
\ifpdf
  \DeclareGraphicsExtensions{.eps,.pdf,.png,.jpg}
\else
  \DeclareGraphicsExtensions{.eps}
\fi

\usepackage{enumitem}
\setlist[enumerate]{leftmargin=.5in}
\setlist[itemize]{leftmargin=.5in}

\newsiamremark{remark}{Remark}
\newsiamremark{hypothesis}{Hypothesis}
\crefname{hypothesis}{Hypothesis}{Hypotheses}
\newsiamthm{claim}{Claim}
\newsiamremark{fact}{Fact}
\crefname{fact}{Fact}{Facts}

\headers{Scalable methods for HMM and fHMM}{R.~Barrios, I.~Sgouralis}

\title{Tensorized algorithms and scalable filtering methods for hidden Markov and factorial hidden Markov models}

\author{Roxana Barrios\thanks{Department of Mathematics, University of Tennessee, Knoxville, TN
  (\email{rbarrio1@vols.utk.edu}).}
\and Ioannis Sgouralis\thanks{Department of Mathematics, University of Tennessee, Knoxville, TN (\email{isgoural@utk.edu}).}}
\usepackage{amsopn}

\usepackage{float}
\DeclareMathOperator{\vect}{vect}

\DeclareMathOperator{\argmax}{argmax}

\usepackage{amsmath,amsfonts,amssymb,mathtools}
\allowdisplaybreaks[1]

\usepackage{cleveref}

\crefname{remark}{remark}{remarks}
\Crefname{remark}{Remark}{Remarks}

\begin{document}

\maketitle

\begin{abstract}
A common method for the representation and analysis of time-series data is the hidden Markov model (HMM), where each observation is associated with a hidden state that evolves over time. However, many real-world systems are influenced by multiple independent factors, which are more naturally represented by factorial hidden Markov models (fHMM), where several hidden Markov chains jointly generate the observed data. Although an fHMM provides a richer and more realistic representation of many real-world systems, it can be reformulated as an equivalent HMM, but with a significantly larger state-space, leading to a severe increase in computational cost. In particular, the forward filtering algorithm, which is central to evaluation, decoding, and estimation tasks, becomes prohibitively expensive even for small systems. This work focuses on developing scalable methods for time-series analysis using tensor algebra to exploit the multidimensional structure of fHMM directly, without constructing intermediate HMM representations. Our novel filtering approach significantly improves computational performance and enables the efficient analysis of large systems and datasets, extending the scope of fHMM and providing a practical framework for data intensive applications.
\end{abstract}


\section{Introduction}

Data used in scientific research and technology may exhibit varying statistical properties throughout their course of acquisition, requiring dynamical models for their analysis. 
Such models are mathematical representations developed using first- or phenomenological principles that describe how the data statistics evolve over time. Examples in which dynamical models are extensively used include weather predictions, climate measurements, biological and physiological assays, financial market data, and molecular dynamics in physics, chemistry, and biology. Today, dynamical models for data analysis are crucial to forming reliable workflows and machine learning pipelines that until recently have been limited to static models mainly due to their simplicity and lower computational needs.

The prototypical example of time-dependent data is the time-series. A time-series is a sequence of datapoints or empirical measurements acquired at specific times arranged in chronological order. The analysis of such data most often relies on dedicated dynamical models that are specialized to the problem and system at hand. Nevertheless, because of noise, stochastic behavior of the data generation, hidden effects, system specific data acquisition properties, or a combination of these, time-series analysis with generic dynamical models remains a challenging task. 

A common approach to time-series analysis with a wide range of applications is offered by hidden Markov models (HMM), where each observation in a given dataset is associated with a hidden state that evolves over time following Markovian dynamics \cite{rabiner,bishop}. HMMs provide a general probabilistic framework to describe how data statistics evolve through initial and transition state distributions that can be customized to match a broad spectrum of modeling scenarios.

Nevertheless, in many real-world scenarios, the statistics of the observed data are influenced by multiple independent factors. Applications include speech and audio processing \cite{ghahramanifhmms,durrieu2013sffhmm, wohlmayr2011multipitch}, computer vision \cite{wohlmayr2011multipitch, paeng2021visual}, natural language processing \cite{ramanujam2009supertagging, li2011pronoun}, bioinformatics \cite{bioinformatics}, genomics \cite{barangenetic, cancerfhmm, liu2018calis}, neuroscience \cite{gong2024fslds}, smart energy systems \cite{liu2024iobfhmm}, financial time-series analysis \cite{saidane2022fhmv, augustyniak2019fhmv}, activity recognition \cite{kumar2023nilm}, robotics \cite{orlovsavko2022famm}, fault detection, energy systems \cite{yan2022efhmm}, and single molecule experiments \cite{sgouralis_ihmm,sgouralis_blFRET}, among others. For such cases, factorial hidden Markov models (fHMM), where multiple hidden Markov chains instead of a single one may jointly influence the evolving data statistics, offer a more appropriate mathematical representation \cite{ghahramanifhmms,sgouralis_book} while maintaining the modeling flexibility and customization of HMM.

With a dynamical model such as an HMM or fHMM, time-series analysis leads to the investigation of mainly three types of problems \cite{rabiner,bishop,sgouralis_book}. These problems are defined as follows:
\begin{itemize}
\item \textbf{Evaluation problem:} Given values for the model parameters, compute the likelihood of the observed data. This problem is solved by the \emph{forward filtering algorithm}.

\item \textbf{Decoding problem:} Given values for the model parameters, compute the most likely state sequence. This problem is solved by the \emph{Viterbi algorithm}.

\item \textbf{Estimation problem:} Estimate the values of the model parameters. This problem is solved by \emph{specialized Monte Carlo procedures}.
\end{itemize}
All three problems require the application of forward filtering \cite{bishop,max_little,sgouralis_book,rabiner} alone, as for evaluation, or in combination with additional optimization or sampling stages, as in decoding and estimation, respectively. For this reason, filtering algorithms are of paramount importance in time-series analysis \cite{rabiner,sgouralis_book}.

Although fHMMs provide a richer modeling framework, as we show herewith, they can be reformulated as equivalent HMMs with \emph{considerably larger} state-spaces than their original factorial subsystems. As a result, standard filtering algorithms developed for HMMs become computationally expensive when invoked in fHMMs. This limitation restricted the application of fHMMs to relatively small systems that are limited in either the number or sizes of their subsystems.

The present study focuses on developing new filtering and related algorithms adapted to fHMM that scale well with multiple subsystems and large state-spaces. Our approach relies on multilinear algebra and exploits the multidimensional structure of the factorial model dynamics in an fHMM directly, without constructing intermediate HMM representations. This, in combination with the recent improvements in computational packages that implement highly optimized support for tensor operations, such as NumPy \cite{numpy} and MATLAB \cite{matlab}, leads to significantly improved performance, allowing fHMMs to become workhorses for machine learning applications and time-series analysis.

Our study is organized as follows. In \cref{sec:background} we summarize the necessary background on multilinear and tensor algebra and the mathematical formulation of HMM and fHMM. In \cref{sec:main} we present a unified review of existing filtering schemes for HMM and present our novel tensorized algorithms that are adapted to fHMM. In \cref{sec:results} we present numerical results that demonstrate the validity and performance of our novel algorithmic schemes, as well as head-to-head comparisons with their naive counterparts. Conclusions and remarks for further development follow in
\cref{sec:conclusions}. We provide detailed derivations and proofs of the formulas presented herewith in the \textsc{supplementary materials.}

\section{Background}
\label{sec:background}

In this section, we summarize the definitions and introduce the notation used in the subsequent sections. We begin with the necessary background on multilinear and tensor algebra and conclude with the formulation of HMM and fHMM.

\subsection{Tensors}

In this work, it is sufficient to consider a \emph{tensor} as a regularly arranged and fully populated multidimensional array formed by elements taken from a specified field such as the real or complex numbers \cite{delathauwer,koldatensor}. Such an array is characterized by an \emph{order} or \emph{dimensionality} that determines the number of its \emph{directions} or \emph{modes} or \emph{ways}. In turn, the directions of a tensor are characterized by their \emph{dimensions} or \emph{sizes}. The order, directions, and dimensions of a tensor are collectively termed \emph{shape.}

More specifically, we denote a tensor $\mathbb{T}$ on a field $\mathbb{F}$ of order $K$ and dimensions $I^{1:K}$ with 
\begin{align*}
\mathbb{T}\in\mathbb{F}^{I^1\times I^2\times\cdots\times I^K}
.
\end{align*}
This tensor consists of $I=I^1I^2\cdots I^K$ regularly arranged elements that are spread in the directions $1,2,\dots,K$. The individual \emph{elements} of $\mathbb{T}$ are indicated with $\mathbb{T}_{i^1,i^2,\dots,i^K}$ where the \emph{subscript} or \emph{index} $i^k$ associated with the $k$\textsuperscript{th} direction takes values in $\{1,2,\ldots,I^k\}$.

For example, $\mathbb{T}\in\mathbb{R}^{2\times3\times5}$ denotes a tensor on the real numbers of order $K=3$; directions $1,2,3$; and dimensions $I^1=2,I^2=3,I^3=5$. In total, this tensor consists of $I=30$ elements.

Low order tensors, along with their common names and those of their directions, include:
\begin{itemize}

\item Order 1: these tensors are \emph{vectors,} e.g., $\mathcal{T}\in \mathbb{F}^{I^1}$, with elements spread over the column-direction, i.e., direction 1.

\item Order 2: these tensors are \emph{matrices,} e.g., $T\in \mathbb{F}^{I^1\times I^2}$, with elements spread over the column- and row-direction, i.e., directions 1 and 2, respectively.

\item Order 3: these tensors are \emph{books} or \emph{stacks}, e.g., $\mathbb{T}\in\mathbb{F}^{I^1\times I^2\times I^3}$, with elements spread over the column-, row-, and page- or tube- or channel-direction, i.e., directions 1, 2, and 3, respectively.

\end{itemize}

\subsubsection{Basic operations}

For any tensor $\mathbb{T}\in\mathbb{F}^{I^1\times I^2\times\cdots\times I^K}$, we denote the \emph{total sum} of its elements by 
\begin{align*}
[\mathbb{T}]=\sum_{i^K}\cdots\sum_{i^2}\sum_{ i^1}\mathbb{T}_{i^1,i^2,\dots,i^K}
.
\end{align*}
For two tensors $\mathbb{T}',\mathbb{T}''$ of the same shape, we denote their \emph{Hadamard product} by $\mathbb{T}'\odot\mathbb{T}''$. This is a new tensor of the same shape as the original ones whose elements are obtained by element-wise products of the original elements.

For scalars $\tau\in\mathbb{F}$, the \emph{multiplication} of the scalar and a tensor $\mathbb{T}$ is denoted with $\tau\mathbb{T}$ and defined element-wise, resulting in a new tensor shaped similarly to the original one. Multiplication by scalar is naturally extended to \emph{division} by scalar.

\subsubsection{Reshaping operations}

The elements of a tensor can be re-arranged to form new tensors. Such re-arrangement of elements is termed \emph{reshaping} and may or may not change the order of the original tensor.

Re-arrangement of elements that change a tensor's order may take one of two forms:
\emph{folding,} where an original tensor yields a higher-order tensor with the same elements; and \emph{unfolding,} where an original tensor yields a lower-order tensor with the same elements \cite{koldatensor}.

In this study, we consider only the following pair of folding/unfolding operations:

\textbullet~The \emph{vectorization,} which is an unfolding operation, is denoted by
\begin{align*}
\vect^{+1}:\mathbb{F}^{I^1\times I^2\times\cdots\times I^K} \mapsto \mathbb{F}^I
\end{align*}
where $I=I^1I^2\cdots I^K$. It is a non-degenerate linear transformation that transforms any tensor $\mathbb{T}$ into a vector $\mathcal{T}$ of conforming dimension. In this transformation, the elements $\mathcal{T}_i$ of the resulting vector and the elements $\mathbb{T}_{i^1,i^2,\dots,i^K}$ of the original tensor are related by
\begin{align}
\label{eq:revlex_1}
i&=i^1+\sum_{k=2}^K\left(i^k-1\right)\prod_{k'=1}^{k-1}I^{k'}
.
\end{align}

\textbullet~The \emph{inverse vectorization,} which is a folding operation, is denoted by
\begin{align*}
\vect^{-1} :\mathbb{F}^I\mapsto\mathbb{F}^{I^1\times I^2\times\cdots\times I^K}
\end{align*}
where $I=I^1I^2\cdots I^K$. This is also a non-degenerate linear transformation that transforms a vector $\mathcal{T}$ into a tensor $\mathbb{T}$ of conforming shape. In this transformation, the elements $\mathbb{T}_{i^1,i^2,\dots,i^K}$ of the resulting tensor and the elements $\mathcal{T}_i$ of the original vector are related by
\begin{align}
\label{eq:revlex_2}
i^k&= 1+\left\lfloor\frac{i-1}{\prod_{k'=1}^{k-1}I^{k'}}\right\rfloor
\left(\bmod\, I^k\right)
,&k&=1,\dots,K.
\end{align}

With these definitions, $\vect^{+1}$ and $\vect^{-1}$ form a pair of \emph{mutually inverse operators} as they satisfy \begin{align}
\label{eq:vect_id}
\vect^{-1}\left(\vect^{+1}(\mathbb{T})\right)&=\mathbb{T}
,&
\vect^{+1}\left(\vect^{-1}(\mathcal{T})\right)&=\mathcal{T}
,
\end{align}
for any tensor $\mathbb{T}$ and conforming vector $\mathcal{T}$. Furthermore, the pair is compatible with total sums, multiplications by scalars, and Hadamard products as they satisfy
{\small
\begin{align}
\label{eq:vec_sum_tensor}
[\mathbb{T}]&=[\vect^{+1}(\mathbb{T})],
&
[\mathcal{T}]&=[\vect^{-1}(\mathcal{T})],
\\
\vect^{+1}(\tau\mathbb{T})&=\tau\vect^{+1}(\mathbb{T}),
&
\vect^{-1}(\tau\mathcal{T})&=\tau\vect^{-1}(\mathcal{T}),
\label{eq:vec_Hadamard_scalar}
\\
\vect^{+1}(\mathbb{T}'\odot\mathbb{T}'')
&=\vect^{+1}(\mathbb{T}')\odot \vect^{+1}(\mathbb{T}''),
&
\label{eq:vec_Hadamard}
\vect^{-1}(\mathcal{T}'\odot\mathcal{T}'')
&=\vect^{-1}(\mathcal{T}')\odot \vect^{-1}(\mathcal{T}''),
\end{align} }
for any scalar $\tau$ and conforming tensors $\mathbb{T},\mathbb{T}',\mathbb{T}''$ and vectors $\mathcal{T},\mathcal{T}',\mathcal{T}''$.

\begin{remark}
\label[remark]{rem:ass}
\Cref{eq:revlex_1,eq:revlex_2}, that encode the correspondence of subscripts $i\leftrightarrow(i^1,i^2,\dots,i^K)$ for both operators, follow the \emph{column-major convention}. For example, for tensors of order $K=3$ and dimensions $I^1=2,I^2=3,I^3=5$, the vector and tensor subscripts are tabulated in \cref{tab:vect}. As can be seen, the first index, i.e.,~$i^1$, is the fastest; while the last index, i.e.,~$i^K$, is the slowest one.
\end{remark}

\begin{table}[tbhp]
\centering
\caption{Correspondence of subscripts $i\leftrightarrow(i^1,i^2,i^3)$ used in $\vect^{+1}$ and $\vect^{-1}$ for tensors in $\mathbb{F}^{2\times3\times5}$.}
\label{tab:vect}
\begin{tabular}{cc}
$i$ &$(i^1,i^2,i^3)$\\
\hline\hline
$1$   &$(1,1,1)$    \\
$2$   &$(2,1,1)$    \\
$3$   &$(1,2,1)$    \\
$4$   &$(2,2,1)$    \\
$5$   &$(1,3,1)$    \\
$6$   &$(2,3,1)$    \\
$7$   &$(1,1,2)$    \\
$8$   &$(2,1,2)$    \\
$9$   &$(1,2,2)$    \\
$10$  &$(2,2,2)$    \\
\hline\hline
\end{tabular}\qquad\begin{tabular}{cc}
$i$ &$(i^1,i^2,i^3)$\\
\hline\hline
$11$  &$(1,3,2)$   \\
$12$  &$(2,3,2)$   \\
$13$  &$(1,1,3)$   \\
$14$  &$(2,1,3)$   \\
$15$  &$(1,2,3)$   \\
$16$  &$(2,2,3)$   \\
$17$  &$(1,3,3)$   \\
$18$  &$(2,3,3)$   \\
$19$  &$(1,1,4)$   \\
$20$  &$(2,1,4)$   \\
\hline\hline
\end{tabular}\qquad\begin{tabular}{cc}
$i$ &$(i^1,i^2,i^3)$\\
\hline\hline
$21$  &$(1,2,4)$   \\
$21$  &$(2,2,4)$   \\
$23$  &$(1,3,4)$   \\
$24$  &$(2,3,4)$   \\
$25$  &$(1,1,5)$   \\
$26$  &$(2,1,5)$   \\
$27$  &$(1,2,5)$   \\
$28$  &$(2,2,5)$   \\
$29$  &$(1,3,5)$   \\
$30$  &$(2,3,5)$   \\
\hline\hline
\end{tabular}
\end{table}

\subsubsection{Composition operations}

Tensors of conforming shapes can be combined to form new tensors \cite{delathauwer,koldatensor}. In this study, we consider four such operations:

\textbullet~The \emph{mode-$k$ stretch} is defined between a tensor and a vector of conforming dimension. Specifically, for
\begin{align*}
\mathbb{A}&\in\mathbb{F}^{I^1\times\cdots\times I^{k-1}\times I^k\times I^{k+1}\times\cdots\times I^K}
\\
\mathcal{B}&\in\mathbb{F}^{I^k}
\end{align*}
the mode-$k$ stretch is denoted by $\mathbb{A}\odot^k\mathcal{B}=\mathbb{T}$. This is a new tensor
\begin{align*}
\mathbb{T}\in\mathbb{F}^{I^1\times \cdots \times I^{k-1}\times I^k\times I^{k+1}\times \cdots \times I^K}
\end{align*}
of the same shape as the original one and its elements are given by
\begin{equation*}
\mathbb{T}_{i^1,\ldots,i^{k-1},i,i^{k+1},\ldots,i^K} 
=
\mathbb{A}_{i^1,\ldots,i^{k-1},i,i^{k+1},\ldots,i^K} 
\mathcal{B}_i
.
\end{equation*}

\textbullet~The \emph{mode-$k$ product} is defined between a tensor and a matrix of conforming dimensions. Specifically, for
\begin{align*}
\mathbb{A}&\in\mathbb{F}^{I^1\times\cdots\times I^{k-1}\times I^k\times I^{k+1}\times\cdots\times I^K}
\\
B&\in\mathbb{F}^{J\times I^k}
\end{align*}
the mode-$k$ product is denoted by $\mathbb{A}\times^kB=\mathbb{T}$. This is a new tensor
\begin{align*}
\mathbb{T}\in\mathbb{F}^{I^1\times \cdots \times I^{k-1}\times J\times I^{k+1}\times \cdots \times I^K}
\end{align*}
of the same order as the original one and its elements are given by
\begin{equation*}
\mathbb{T}_{i^1,\ldots,i^{k-1},j,i^{k+1},\ldots,i^K } 
=\sum_{i}
\mathbb{A}_{i^1,\ldots,i^{k-1},i,i^{k+1},\ldots,i^K } 
B_{j,i}
.
\end{equation*}

\textbullet~The \emph{modes-$k,\ell$ contraction} is defined between two tensors of conforming dimensions. Specifically, for
\begin{align*}
\mathbb{A}&\in\mathbb{F}^{I^1\times\cdots\times I^{k-1}\times I^k\times I^{k+1}\times\cdots\times I^K}
\\
\mathbb{B}&\in\mathbb{F}^{J^1\times\cdots\times J^{\ell-1}\times I^k\times J^{\ell+1}\times\cdots\times J^L}
\end{align*}
the modes-$k,\ell$ contraction is denoted by $\mathbb{A}\times^{k,\ell}\mathbb{B}=\mathbb{T}$. This is a new tensor 
\begin{align*}
\mathbb{T}\in\mathbb{F}^{I^1\times\cdots\times I^{k-1}\times I^{k+1}\times\cdots\times I^K\times J^1\times\cdots\times J^{\ell-1}\times J^{\ell+1}\times\cdots\times J^L}
\end{align*}
of order larger than the orders of the original ones and its elements are given by
\begin{align*}
\mathbb{T}_{i^1,\ldots,i^{k-1},i^{k+1},\ldots,i^K,j^1,\ldots,j^{\ell-1},j^{\ell+1},\ldots,j^L} 
=\sum_{i}
\mathbb{A}_{i^1,\ldots,i^{k-1},i,i^{k+1},\ldots,i^K} 
\mathbb{B}_{j^1,\ldots,j^{\ell-1},i,j^{\ell+1},\ldots,j^L} 
.
\end{align*}

\textbullet~The \emph{outer product} is defined between two tensors of arbitrary shape. Specifically, for
\begin{align*}
\mathbb{A}&\in\mathbb{F}^{I^1\times\cdots\times I^K}
\\
\mathbb{B}&\in\mathbb{F}^{J^1\times\cdots\times J^L}
\end{align*}
the outer product is denoted by $\mathbb{A}\times\mathbb{B}=\mathbb{T}$. This is a new tensor 
\begin{align*}
\mathbb{T}\in\mathbb{F}^{I^1\times\cdots\times I^K\times J^1\times\cdots\times J^L}
\end{align*}
of order larger than the orders of the original ones and its elements are given by
\begin{align*}
\mathbb{T}_{i^1,\ldots,i^K,j^1,\ldots,j^L} 
=
\mathbb{A}_{i^1,\ldots,i^K} 
\mathbb{B}_{j^1,\ldots,j^L}
.
\end{align*}

\subsection{Special low order operations}

In this study, we also use the following special operations that apply only on low order tensors, specifically vectors and matrices.

\textbullet~The ordinary \emph{matrix-vector product} is defined between a matrix and a vector of conforming dimensions. Specifically, for $
A\in \mathbb{F}^{I\times J}$ and $\mathcal{B}\in \mathbb{F}^{J}$
the \emph{matrix-vector product} is denoted by $ A\cdot \mathcal{B} $. This is a new vector $\mathcal{C}\in \mathbb{F}^{I}$ and its elements are given by
\begin{align*}
\mathcal{C}_i
=
\sum_{j=1}^{J}
A_{ij}\mathcal{B}_j
.
\end{align*}

\textbullet~The \emph{Kronecker product} is defined between vectors and matrices. Specifically, for vectors  $\mathcal{A}\in\mathbb{F}^{I}$ and $\mathcal{B}\in\mathbb{F}^{J}$ 
the \emph{Kronecker product} is denoted by $\mathcal{A} \otimes \mathcal{B}$. This is a new vector 
$\mathcal{C}\in \mathbb{F}^{IJ}$ and is given blockwise by
\begin{align*}
\mathcal{C}
=
\begin{bmatrix}
\mathcal{A}_{1} \mathcal{B} \\
\mathcal{A}_{2} \mathcal{B} \\
\vdots \\
\mathcal{A}_{I}\mathcal{B}
\end{bmatrix}
.
\end{align*}
Similarly, for matrices  $A\in \mathbb{F}^{I^1\times I^2}$ and $B\in\mathbb{F}^{J^1\times J^2}$
the \emph{Kronecker product} is denoted by $A \otimes B$. This is a new matrix
$C\in\mathbb{F}^{I^{1}J^1\times I^{2}J^2}$ and is given blockwise by
\begin{align*}
C
=
\begin{bmatrix}
A_{11} B & A_{12} B & \cdots & A_{1I^2} B \\
A_{21} B & A_{22} B & \cdots & A_{2I^2} B \\
\vdots & \vdots & \ddots & \vdots \\
A_{I^11} B & A_{I^12} B & \cdots & A_{I^1I^2} B
\end{bmatrix}
.
\end{align*}

\subsection{Successive operations}
\label{sec:trains_tens}

Trains of successive Kronecker products, developed between vectors $\mathcal{T}_k\in\mathbb{F}^{I^k}$, satisfy 
\begin{align}
\label{eq:initial_kron_to_outer}
\mathcal{T}_K\otimes \cdots\otimes\mathcal{T}_1
&=
\vect^{+1}\left(\mathcal{T}_1\times \cdots \times\mathcal{T}_K\right)
.
\end{align}
Trains of successive mode-$k$ stretches, developed between vectors $\mathcal{T}_k\in \mathbb{F}^{I^k}$ and a tensor $\mathbb{T}\in\mathbb{F}^{I^1\times\cdots\times I^K}$ of conforming dimensions, satisfy.
\begin{align}
\label{train_modek_stretch}
\mathbb{T}\odot
\left(\mathcal{T}_1\times\cdots\times \mathcal{T}_K\right)
=
\mathbb{T}\odot^1\mathcal{T}_1\cdots\odot^K\mathcal{T}_K.
\end{align}
Trains of successive Kronecker products, developed between matrices $T_k\in \mathbb{F}^{J^k\times I^k}$ and a vector $\mathcal{T}\in\mathbb{F}^{I^1\cdots I^K}$ of conforming dimension, satisfy 
\begin{align}
\label{tensor_vect2}
\left( T_K\otimes \cdots \otimes T_1\right)\cdot \mathcal{T}
&=
\vect^{+1}\left(\vect^{-1}(\mathcal{T})\times^1 T_1 \cdots \times^K T_K\right).
\end{align}
Trains of successive mode-$k$ products, developed between a tensor $\mathbb{T}\in\mathbb{F}^{I^1\times\cdots\times I^K}$ and matrices $T_k\in \mathbb{F}^{J^k\times I^k}$ of conforming dimensions, satisfy 
\begin{align}
\label{tensor_cont2}
\mathbb{T}\times^1 T_1\cdots \times^K T_K
&=
\mathbb{T}\times^{1,2} T_1\cdots \times^{1,2} T_K
.
\end{align}
Detailed proofs of \cref{eq:initial_kron_to_outer,train_modek_stretch,tensor_vect2,tensor_cont2} are given in the \textsc{supplementary materials}.

\subsection{Models of time-series data}

We now review the basic definitions of the dynamical models used in conjunction with time-series data that are relevant to this study.

\subsubsection{The hidden Markov model}\label{sec:basic_HMM}

A \emph{hidden Markov model} (HMM) is a mathematical model that describes probabilistic relationships developed over time between noisy observations and dynamically changing latent states \cite{rabiner,bishop,sgouralis_book}. In an HMM the observations take the form of time-series. Each observation $w_n$ in the series is associated with a hidden state $s_n$ that can change over time.

Time in an HMM is modeled sequentially and, in our notation, is denoted by indices
\begin{align*}
n=1,\dots,N
\end{align*}
with higher indices indicating observations made chronologically later than lower ones and vise versa lower indices indicating observations made chronologically earlier than higher ones.

To model dynamics in an HMM, it is useful to distinguish between passing states and constitutive states. The \emph{passing states} are the random variables $s_n$ that model the hidden states over time. In contrast, the \emph{constitutive states} are a fixed set of abstract values $\sigma_m$ that each $s_n$ can take. In this study, we use $
m=1,\dots,M$ to index constitutive states and refer to 
\begin{align*}
\mathbb{S}=\{\sigma_m\}_{m=1}^M
\end{align*}
as the \emph{state-space} of the HMM. 

An HMM is described in terms of probability distributions. A fundamental assumption is that these distributions satisfy the Markov property, which reads
\begin{align}\label{eq:mark_HMM}
    p\left(s_n| s_{n-1}, s_{n-2}, \ldots, s_1\right)&=p\left(s_n| s_{n-1}\right)
    ,&
    n&=2,\dots,N.
\end{align}
Due to the Markov property, the transition dynamics in an HMM are determined by the \emph{transition probabilities} $p\left(s_n| s_{n-1}\right)$ between successive states. These are denoted with
\begin{align*}
\pi_{s_{n-1}\to s_n}&=p\left(s_n| s_{n-1}\right)
,&
n&=2,\dots,N
\end{align*}
and, considered over the entire state-space, tabulated into a transition probability matrix
\begin{align*}
\Pi =
\begin{bmatrix}
\pi_{\sigma_1 \to \sigma_1} & \pi_{\sigma_2 \to \sigma_1} & \cdots & \pi_{\sigma_M \to \sigma_1} \\
\pi_{\sigma_1 \to \sigma_2} & \pi_{\sigma_2 \to \sigma_2} & \cdots & \pi_{\sigma_M \to \sigma_2} \\
\vdots & \vdots & \ddots & \vdots \\
\pi_{\sigma_1 \to \sigma_M} & \pi_{\sigma_2 \to \sigma_M} & \cdots & \pi_{\sigma_M \to \sigma_M}
\end{bmatrix}
\end{align*}
which contains the probabilities of transitioning from any $\sigma$ to any $\sigma'$ within $\mathbb{S}$. Each \emph{column} of $\Pi$ corresponds to a departing state, and each row corresponds to a possible arriving state. Because its elements are \emph{conditional probabilities,} the matrix $\Pi$ is, by its definition, normalized such that
\begin{align*}
\sum_{\sigma'\in\mathbb{S}} \pi_{\sigma\to \sigma'} &= 1
,&
\sigma&\in\mathbb{S}.
\end{align*}
Due to the normalization condition, which applies to each column separately, the transition matrix is often broken down into individual \emph{transition probability vectors}
\begin{align*}
\Pi=\{\pi_{\sigma}\}_{\sigma\in\mathbb{S}}
\end{align*}
that we denote with $\pi_\sigma$. These are defined by
\begin{align*}
\pi_\sigma&=
\begin{bmatrix}
\pi_{\sigma\to\sigma_1}
\\
\pi_{\sigma\to\sigma_2}
\\
\vdots
\\
\pi_{\sigma\to\sigma_M}
\end{bmatrix}
,&
\sigma&\in\mathbb{S}.
\end{align*}

Because the Markov property of \cref{eq:mark_HMM} does not apply on $s_1$, an HMM also requires the specification of \emph{initial probabilities,} which are denoted with
\begin{align*}
\rho_{s_1}&=p(s_1)
.
\end{align*}
Similarly to the transition probabilities, the initial probabilities are normalized such that
\begin{align*}
\sum_{\sigma\in\mathbb{S}} \rho_\sigma &= 1
\end{align*}
and, considered over the entire state-space, gathered into an initial probability vector
\begin{align*}
\rho&=
\begin{bmatrix}
\rho_{\sigma_1}
\\
\rho_{\sigma_2}
\\
\vdots
\\
\rho_{\sigma_M}
\end{bmatrix}
.
\end{align*}

Finally, measurements in an HMM are modeled by the distributions $p(w_n|s_n)$ which are problem specific. These are termed emission distributions, and we denote them with
\begin{align*}
G_{\phi_{s_n}}^{\psi}(w_n)&=p(w_n|s_n)
,&
n&= 1,\dots,N.
\end{align*}
The emission distributions model the statistics of the observed datapoints $w_n$ and link them to their corresponding hidden states $s_n$. The linking relationship depends on \emph{emission parameters} such as $\phi_{s_n}$, that differ among states, and also parameters such as $\psi$, that may be common to all states. To simplify the notation, we gather the former together in
\begin{align*}
\phi=\{\phi_\sigma\}_{\sigma\in\mathbb{S}}
.
\end{align*}

\subsubsection{The factorial hidden Markov model}
\label{sec:basic_fHMM}

The HMM, described above, assumes a single Markov chain that drives the dynamics of the time-series under analysis. However, complicated modeling scenarios, where observations are influenced by \emph{multiple} and \emph{independent} Markov chains, are also of practical interest \cite{ghahramanifhmms,sgouralis_ihmm}.
For example, speech recognition \cite{ghahramanifhmms} naturally gives rise to dynamical models with multiple latent chains, since the observed signal is often better explained by several underlying processes (e.g., phonetic content, linguistic structure, and acoustic/channel effects) evolving in parallel rather than by a single one. In addition, in experimental biophysics and specifically in the analysis of raw measurements from single-molecule F{\"o}rster resonance energy transfer (smFRET) experiments \cite{lakowicz}, the observed fluorescence intensities are influenced not only by the conformation of the molecule, but also by the photo-physics of the donor and acceptor fluorophores, which may stochastically transition between bright and dark independently of the molecule \cite{sgouralis_blFRET,sgouralis_ihmm,sgouralis_icon}. In these cases, the modeled time-series arises from the combined effect of several independent stochastic processes that evolve simultaneously.
To account for such scenarios, the factorial hidden Markov model, which we describe next, has been developed as a natural extension of the HMM.

A factorial hidden Markov model (fHMM) is a generalization of the HMM that models multiple independent sequences of hidden states that are jointly related to the same observed data \cite{ghahramanifhmms}. This is applicable in modeling cases where there are multiple subsystems contributing simultaneously to a time-series; for example, in smFRET the fHMM can model the conformational states of the molecule and also the photo-physical states of the two fluorophores. With an fHMM, it is possible to separate these processes since each is associated with its own hidden state, state-space, and dynamics. This type of decomposition helps to reduce data misinterpretation and analysis bias, as the whole model remains physically faithful rather than with an HMM.

More precisely, a fHMM models multiple \emph{subsystems} that, in this study, we index with \emph{superscripts} $k=1,\dots,K$. Each subsystem has its own state-space of its own size and constitutive states that we denote with
\begin{align*}
\mathbb{S}^{k}=\left\{\sigma^{k}_{m^k}\right\}_{m^k=1}^{M^{k}}
.
\end{align*}
In addition, each subsystem has its own passing states $s^{k}_n$ that satisfy the Markov property
\begin{align}\label{eq:mark_fHMM}
    p\left(s_n^{k}| s_{n-1}^{k}, s_{n-2}^{k}, \ldots, s_1^{k}\right)&=p\left(s_n^{k}| s_{n-1}^{k}\right)
    ,&
    n&=2,\dots,N.
\end{align}
Because of the Markov property, each subsystem has its own transition probabilities 
\begin{align*}
\pi^k_{s^{k}_{n-1}\to s_n^{k}}=p\left(s_n^{k}| s^{k}_{n-1}\right)
\end{align*}
which are tabulated into separate transition probability matrices
\begin{align*}
\Pi^{k}
&=
\left\{\pi^{k}_{\sigma^{k}}\right\}_{\sigma^{k}\in\mathbb{S}^{k}}
\end{align*}
that are formed by each subsystem's transition probability vectors. Furthermore, each subsystem has its own initial probabilities
\begin{align*}
\rho^k_{s_1^k}
=
p\left(s_1^{k}\right)
\end{align*}
that are gathered into initial probability vectors $\rho^{k}$.

Finally, measurements in an fHMM are modeled by the emission distributions $p\left(w_n|s_n^{1:K}\right)$ that may depend on \emph{all} subsystems. These are given by
\begin{align*}
G^\psi_{
\phi^{1}_{s^{1}_n}
,
\dots,
\phi^{K}_{s^{K}_n}
}
(w_n)
&=
p\left(w_n|s_n^{1:K}\right)
,&
n&= 1,\dots,N,
\end{align*}
where each subsystem has its own emission parameters which, to simplify the notation, we gather into
\begin{align*}
\phi^{k}
=
\left\{
\phi^{k}_{\sigma^{k}}
\right\}_{\sigma^{k}
\in\mathbb{S}^{k}}
.
\end{align*}

\subsubsection{Equivalence of HMM and fHMM}
\label{sec:fHMM_to_HMM_conv}

It can be easily seen that an HMM is equivalent to an fHMM consisting of just one subsystem. However, as we now show, an fHMM is \emph{also} equivalent to an HMM. Because this equivalence is essential for the development of the filtering schemes in the following sections, below we explain in detail the conventions underlying the state-space and dynamics matching between the two formalisms. 

\paragraph{States}

The fHMM with the notation of \cref{sec:basic_fHMM} can be converted to an HMM with the notation of \cref{sec:basic_HMM}. For this conversion, the passing states of the fHMM are defined by $K$-tuples of the form
\begin{align*}
s_n&=\left(s^1_n,\dots,s^K_n\right),
&
n&=1,\dots,N
.
\end{align*}
Accordingly, the constitutive states of the fHMM are also given by $K$-tuples of the form
\begin{align}
\label{eq:conv_sigmas}
\sigma_m=\left(\sigma_{m^1}^1,\dots, \sigma_{m^K}^K\right)
,
\end{align}
and the state-space is given by the \emph{Cartesian product} of the individual state-spaces
\begin{align}
\mathbb{S}=\mathbb{S}^1\times\cdots\times\mathbb{S}^K
.
\label{eq:cart_S}
\end{align}
Consequently, the size of the resulting state-space is equal to the product
\begin{align}
\label{eq:fHMM_M}
M&=M^1\cdots M^K
\end{align}
formed by the sizes of the original state-spaces.

\begin{remark}
\label[remark]{rem:HMM_fHMM_conv}
As will become apparent in \cref{sec:filtering_fHMM}, the tensorization of the algorithms adapted to fHMM is facilitated provided that the constitutive states $\sigma_m\in\mathbb{S}$ of its equivalent HMM are indexed according to the convention in \cref{rem:ass}. For this reason, in this study we exclusively enumerate the Cartesian product $\mathbb{S}^1\times\cdots\times\mathbb{S}^K$ in \cref{eq:cart_S} by associating the indices $m\leftrightarrow(m^1,\dots,m^K)$ in \cref{eq:conv_sigmas} using the same convention as in \cref{eq:revlex_1,eq:revlex_2}.
\end{remark}

\paragraph{Dynamics}

According to \cref{sec:basic_HMM}, the initial probabilities of the resulting HMM are given by
\begin{align*}
\rho_{\sigma_m}
=
\rho^1_{\sigma^1_{m^1}}
\dots
\rho^K_{\sigma^K_{m^K}}
.
\end{align*}
These are gathered into a column vector that, because of \cref{rem:HMM_fHMM_conv}, is obtained by the Kronecker product
\begin{align}
\label{eq:kron_rho}
\rho=\rho^K\otimes\cdots\otimes\rho^1
.
\end{align}
Similarly, the transition probabilities of the resulting HMM are given by 
\begin{align*}
\pi_{\sigma_m\to\sigma_{m'}}
=
\pi^1_{\sigma^1_{m^1}\to\sigma^1_{m^{K'}}}
\cdots
\pi^K_{\sigma^K_{m^K}\to\sigma^K_{m^{K'}}}
.
\end{align*}
These are gathered into column vectors that, due to \cref{rem:HMM_fHMM_conv}, are also obtained by the Kronecker products
\begin{align*}
\pi_{\sigma_m}=\pi^K_{\sigma_m}\otimes\cdots\otimes\pi^1_{\sigma_m}
.
\end{align*}
In addition, the transition probability vectors are gathered into a square matrix which, because of \cref{rem:HMM_fHMM_conv}, is also obtained by the Kronecker product
\begin{align}
\label{eq:kron_Pi}
\Pi=\Pi^K\otimes\cdots\otimes\Pi^1.
\end{align}

\section{Methods}
\label{sec:main}

Having established our main definitions and notational conventions, we now turn to the development of our algorithmic procedures.

\subsection{Existing vectorized algorithms for HMM}

In this section, we review existing vectorized algorithmic schemes for HMMs that are available in the literature \cite{rabiner,bishop,sgouralis_book}. We focus on the standard forward filtering and backward decoding  recursions that use matrix-vector operations. This will motivate tensor generalizations that we will introduce in subsequent sections.

\subsubsection{Forward filtering for HMM}
\label{sec:filtering_HMM}

The evaluation problem of an HMM, as described in the \textsc{introduction}, requires computation of the \emph{marginal likelihood,} which is given by $p(w_{1:N}|\theta)$ where $\theta=\{\rho,\Pi,\phi,\psi\}$ stands for all the parameters of the HMM. A \emph{naive} way to compute the marginal likelihood may be achieved according to the completion
\begin{align*}
    p(w_{1:N}|\theta)
    &= \sum_{s_1}\sum_{s_2}\cdots\sum_{s_N} p(w_{1:N}, s_{1:N}|\theta)
    \\
    &= \sum_{s_1}
        G_{\phi_{s_1}}^{\psi}(w_1)
        \rho_{s_1}
    \,
    \sum_{s_2}
    G_{\phi_{s_2}}^{\psi}(w_2)        \pi_{s_1\to s_2}
    \cdots
    \,
    \sum_{s_N}
    G_{\phi_{s_N}}^{\psi}(w_N)
    \pi_{s_{N-1}\to s_N}
    .
\end{align*}
Nevertheless, the implementation of this formula is impractical as it requires the evaluation and addition of $M^N$ terms, i.e., $M$ terms for $N$ sums, which is a very computationally demanding task even for small systems \cite{rabiner,sgouralis_book}. 

Instead, an \emph{efficient} way to compute the marginal likelihood $p(w_{1:N}|\theta)$ of a HMM is achieved according to the factorization
\begin{align*}
p\left(w_{1:N}|\theta\right)
&=p(w_1|\theta)\prod_{n=2}^N p\left(w_n|w_{1:n-1},\theta\right)
=\prod_{n=1}^N C_n
\end{align*}
where the factors $C_{1:N}$, depend \emph{implicitly} on the measurements $w_{1:N}$ and the parameters $\theta$ of the model. These factors are defined by
\begin{align*}
C_1&=p(w_1|\theta),
\\
C_n&=p(w_n|w_{1:n-1},\theta)
,& n&=2,\dots,N,
\end{align*}
and computed according to the completions
\begin{align}
\label{eq:HMM_Cn}
C_n&=\sum_{s_n}A'_n(s_n)
,& n&=1,\dots,N
.
\end{align}
A proof of \cref{eq:HMM_Cn} is provided in the \textsc{supplemental materials}. The new terms $\{A_{1:N}'(\sigma)\}_{\sigma\in\mathbb{S}}$ are now defined by
\begin{align}
\label{eq:non_filter1}
A'_1(s_1)&=G^\psi_{\phi_{s_1}}(w_1)\rho_{s_1},
\\
\label{eq: non_filter}
A'_n(s_n)&=G^\psi_{\phi_{s_n}}(w_n)p(s_n|w_{1:n-1},\theta)
,&
n&=2,\ldots,N,
\end{align}
and the numerically robust way of computing them \cite{sgouralis_book,rabiner} is based on the \emph{normalized filters} $\{A_{1:N}(\sigma)\}_{\sigma\in\mathbb{S}}$ which, in turn, are defined by
\begin{align}
\label{eq:filters}
A_n(s_n)&=p(s_n|w_{1:n},\theta),
&
n&=1,\dots,N.
\end{align}
The efficient computation of $\{A_{1:N}(\sigma)\}_{\sigma\in\mathbb{S}}$ is \emph{interlaced} with the computation of $C_{1:N}$ and $\{A'_{1:N}(\sigma)\}_{\sigma\in\mathbb{S}}$. Specifically, the computation of the normalized filters proceeds via
\begin{align}
\label{eq:HMM_An}
A_n(s_n)
&=
\frac{A'_n(s_n)}{C_n}
,&
n&=1,\dots,N
.
\end{align}
For this, $A_1'(s_1)$ is first computed by \cref{eq:non_filter1} and then advanced via
\begin{align}
\label{eq:HHM_An_prime}
A'_n(s_n)
&=
G^\psi_{\phi_{s_n}}(w_n)\sum_{s_{n-1}}\pi_{s_{n-1}\to s_n}A_{n-1}(s_{n-1})
,&n&=2,\dots,N.
\end{align}
Proofs of \cref{eq:HMM_An,eq:HHM_An_prime} are provided in the \textsc{supplemental materials.} 

The implementation of this recursive scheme is simplified by gathering the elements involved $\{G^\psi_{\phi_\sigma}(w_n),A_n'(\sigma),A_n(\sigma)\}_{\sigma\in\mathbb{S}}$ into \emph{vectors} like these
\begin{align*}
\mathcal{G}_n
&=
\begin{bmatrix}
G^\psi_{\phi_{\sigma_1}}(w_n)
\\
\vdots
\\
G^\psi_{\phi_{\sigma_m}}(w_n)
\\
\vdots
\\
G^\psi_{\phi_{\sigma_M}}(w_n)
\end{bmatrix},
&
\mathcal{A}'_n
&=
\begin{bmatrix}
A'_n(\sigma_1)
\\
\vdots
\\
A'_n(\sigma_m)
\\
\vdots
\\
A'_n(\sigma_M)
\end{bmatrix},
&
\mathcal{A}_n
&=
\begin{bmatrix}
A_n(\sigma_1)
\\
\vdots
\\
A_n(\sigma_m)
\\
\vdots
\\
A_n(\sigma_M)
\end{bmatrix}
,&n&=1,\dots,N.
\end{align*}
The recursion then requires $N$ iterations, each consisting of 3 stages, like this
\begin{align*}
\underbrace{
\mathcal{A}'_1\to C_1\to\mathcal{A}_1
}_{\text{1\textsuperscript{st} iteration}}
\quad\to\quad
\underbrace{
\mathcal{A}'_2\to C_2\to \mathcal{A}_2
}_{\text{2\textsuperscript{nd} iteration}}
\quad\to\quad
\underbrace{
\mathcal{A}'_3\to C_3\to\mathcal{A}_3
}_{\text{3\textsuperscript{rd} iteration}}
\quad\to\quad\cdots
.
\end{align*}
In vector format, the \emph{initial iteration,} i.e., $n=1$, proceeds like this
\begin{align}
\mathcal{A}'_1&=\mathcal{G}_1\odot\rho,
\label{eq:ff_hmm_1a}
\\
C_1&=\left[\mathcal{A}'_1\right],
\label{eq:ff_hmm_1b}
\\
\mathcal{A}_1&=\mathcal{A}'_1/C_1.
\label{eq:ff_hmm_1c}
\end{align}
Similarly, \emph{subsequent iterations,} i.e., $n=2,\dots,N$, proceed like this
\begin{align}
\mathcal{A}'_n&=\mathcal{G}_n\odot\left(\Pi\cdot\mathcal{A}_{n-1}\right) ,\label{eq:Anprime}
\\
C_n&=\left[\mathcal{A}'_n\right] ,\label{eq:Cn}
\\
\mathcal{A}_n&=\mathcal{A}'_n/C_n.
\label{eq:An}
\end{align}
The implementation of this recursive scheme allows successive replacement of $\mathcal{G}_n$ with $\mathcal{A}'_n$ and then $\mathcal{A}_n$. This way only one copy of the 3 vectors is required in memory at any given time.

The computational load of each iteration is dominated by the matrix-vector multiplication in \cref{eq:Anprime}. This has an asymptotic cost of $O(M^2)$ per iteration \cite{golub2013matrix}, which leads to an overall cost of the entire recursive scheme of $O(NM^2)$.

\subsubsection{Decoding for HMM}
\label{sec:viterbi_HMM}

The decoding problem of an HMM, as described in the \textsc{introduction}, requires the computation of the \emph{maximizer} $\tilde s_{1:N}$ of $p(s_{1:N}|w_{1:N},\theta)$ where $\theta=\{\rho,\Pi,\phi,\psi\}$ stands for all parameters of the HMM. An efficient way to compute this maximizer is based on the factorization
\begin{align*}
p(s_{1:N}|w_{1:N},\theta)
&=
\left[\prod_{n=1}^{N-1}p(s_n|s_{n+1},w_{1:n},\theta)\right]
p(s_N|w_{1:N},\theta)
\end{align*}
which allows for recursive computation of individual maximizers $\tilde s_n$ starting from $\tilde s_N$ and proceeding sequentially to $\tilde s_1$.

Individual maximizations start based on the equality 
\begin{align*}
p(s_N|w_{1:N},\theta)&=A_N(s_N)
\end{align*}
which is a direct consequence of the definition of the normalized filters in \cref{eq:filters}, and proceed based on the proportionality
\begin{align}
p(s_n|s_{n+1},w_{1:n},\theta)
&\propto
\pi_{s_n\to s_{n+1}}A_n(s_n)
,&n=1,\dots,N-1.
\label{eq:dec_hmm1}
\end{align}
\Cref{eq:dec_hmm1} is proved in the \textsc{supplementary materials.}

In summary, the maximizers are obtained by the Viterbi recursion \cite{sgouralis_book,rabiner} that proceeds like this
\begin{align}
\tilde s_N&=\argmax_\sigma A_N(\sigma)
\\
\tilde s_n&=\argmax_\sigma A_n(\sigma)\pi_{\sigma\to \tilde s_{n+1}}
,&n=1,\dots,N-1.
\end{align}
Computationally, each iteration of the Viterbi algorithm is facilitated by the formation of an \emph{auxiliary vector} $\mathcal{T}_n$ that gathers the objective values under optimization. For the first iteration, this is defined directly through the normalized filter
\begin{align}
\mathcal{T}_N&=
\begin{bmatrix}
A_N(\sigma_1)
\\
\vdots
\\
A_N(\sigma_m)
\\
\vdots
\\
A_N(\sigma_M)
\end{bmatrix}
=\mathcal{A}_N
\end{align}
and for subsequent iterations, this is defined by
\begin{align}
\mathcal{T}_n&=
\begin{bmatrix}
A_n(\sigma_1)\pi_{\sigma_1\to\tilde s_{n+1}}
\\
\vdots
\\
A_n(\sigma_m)\pi_{\sigma_m\to\tilde s_{n+1}}
\\
\vdots
\\
A_n(\sigma_M)\pi_{\sigma_M\to\tilde s_{n+1}}
\end{bmatrix}
=
\mathcal{A}_n
\odot
\mathcal{P}\left(\tilde s_{n+1}\right)
,&n&=1,\dots,N-1,
\end{align}
where the vectors $\{\mathcal{P}\left(\sigma\right)\}_{\sigma\in\mathbb{S}}$ are obtained by
\begin{align}
\mathcal{P}\left(\sigma\right)
&=
\begin{bmatrix}
\pi_{\sigma_1\to\sigma}
\\
\vdots
\\
\pi_{\sigma_m\to\sigma}
\\
\vdots
\\
\pi_{\sigma_M\to\sigma}
\end{bmatrix}
.
\end{align}
The recursion requires $N$ iterations, each consisting of 2 stages, like this
\begin{align*}
\underbrace{
\mathcal{T}_N\to\tilde s_N
}_{\text{1\textsuperscript{st} iteration}}
\quad\to\quad
\underbrace{
\mathcal{T}_{N-1}\to\tilde s_{N-1}
}_{\text{2\textsuperscript{nd} iteration}}
\quad\to\quad
\underbrace{
\mathcal{T}_{N-2}\to\tilde s_{N-2}
}_{\text{3\textsuperscript{rd} iteration}}
\quad\to\quad\cdots
.
\end{align*}
The implementation of this recursive scheme allows successive replacement of $\mathcal{T}_{n}$ with $\mathcal{T}_{n-1}$. In this way, only one copy of all auxiliary vectors is required in memory at any given time.

\subsection{New tensorized algorithms for fHMM}

In this section, we present new tensor based computational algorithms for fHMM that extend beyond the existing literature. Here, based on the equivalence of fHMM and HMM that we showed earlier, we develop filtering schemes that exploit the multidimensional structure of fHMM through tensor operations. Because such operations are nowadays natively supported by highly optimized computer software, our fHMM adapted algorithms allow for improved performance relative to their naive counterparts.

\subsubsection{Forward filtering for fHMM}
\label{sec:filtering_fHMM}

As we showed in \cref{sec:fHMM_to_HMM_conv}, an fHMM is a special case of an HMM. For this reason, the computation of the \emph{marginal likelihood} $p\left(w_{1:N}|\theta\right)$, where $\theta=\{\rho^{1:K}, \Pi^{1:K},\phi^{1:K},\psi\}$ now stands for the parameters of all subsystems, can be computed via the same filtering algorithm as in \cref{sec:filtering_HMM}.

Because of \cref{eq:fHMM_M}, the asymptotic cost of applying HMM's forward filtering algorithm to an fHMM without any modification is given by 
\begin{align}\label{eq:old_complexity}
O\left(NM^2\right)=O\left(NM\prod_{k=1}^KM^k\right).
\end{align}
However, due to \cref{eq:kron_rho,eq:kron_Pi}, which mediate conversion to HMM, a more efficient filtering scheme that is adapted to fHMM can be developed. In this scheme, we consider \emph{adapted filters} and define the tensors 
\begin{align*}
    \mathbb{A}_n &= \vect^{-1}(\mathcal{A}_n)
    ,
\end{align*}
where $\vect^{-1}:\mathbb{R}^M\mapsto\mathbb{R}^{M^1\times\cdots\times M^K}
$ folds the vector filters $\mathcal{A}_n$ of the equivalent HMM to match the dimensions of the subsystems of the fHMM. According to \cref{eq:ff_hmm_1c,eq:An,eq:vec_Hadamard_scalar}, the adapted filters can be obtained by
\begin{align*}
\mathbb{A}_n
= \frac{\mathbb{A}'_n}{C_n} 
\end{align*}
where, by definition, each tensor $\mathbb{A}'_n$ is obtained by
\begin{align*}
\mathbb{A}'_n&=\vect^{-1}(\mathcal{A}_n').
\end{align*}
Now, according to \cref{eq:vec_sum_tensor,eq:ff_hmm_1b,eq:Cn}, the normalizing constants are given by
\begin{align*}
C_n
=[\mathbb{A}_n']
.
\end{align*}
Finally, according to \cref{eq:ff_hmm_1a,eq:kron_rho,eq:initial_kron_to_outer,train_modek_stretch}, $\mathbb{A}_1'$ is obtained by tensor stretches
\begin{align}
\mathbb{A}_1'
=\mathbb{G}_1\odot\left(\rho^1\odot^1\cdots \odot^K \rho^K\right)
,
\label{eq:ini_vector5}
\end{align}
and, similarly, according to \cref{tensor_vect2,tensor_cont2,eq:kron_Pi,eq:Anprime}, subsequent $\mathbb{A}_n'$ are obtained by tensor contractions
\begin{align}
\mathbb{A}_n'
&=\mathbb{G}_n \odot\left( \mathbb{A}_{n-1}\times^{1,2} \Pi^1\cdots \times^{1,2} \Pi^K\right)
,
&n&=2,\dots,N.
\label{eq:ini_vector5n}
\end{align}
Detailed proofs of \cref{eq:ini_vector5,eq:ini_vector5n} are provided in the \textsc{supplementary materials}. Here, by definition, the tensors $\mathbb{G}_n$ are obtained by
\begin{align*}
\mathbb{G}_n&=\vect^{-1}\left(\mathcal{G}_n\right).
\end{align*}
As with the HMM, the implementation of this recursion requires $N$ iterations, each consisting of 3 stages, like this
\begin{align*}
\underbrace{
\mathbb{A}'_1\to C_1\to\mathbb{A}_1
}_{\text{1\textsuperscript{st} iteration}}
\quad\to\quad
\underbrace{
\mathbb{A}'_2\to C_2\to \mathbb{A}_2
}_{\text{2\textsuperscript{nd} iteration}}
\quad\to\quad
\underbrace{
\mathbb{A}'_3\to C_3\to\mathbb{A}_3
}_{\text{3\textsuperscript{rd} iteration}}
\quad\to\quad\cdots
.
\end{align*}
The \emph{initial iteration,} i.e., $n=1$, proceeds like this 
\begin{align}
\mathbb{A}'_1&=\mathbb{G}_1 \odot\left( \rho^1\odot^1 \cdots \odot^K \rho^K\right),
\label{eq:ff_fhmm_1a}
\\
C_1&=\left[\mathbb{A}'_1\right],
\label{eq:ff_fhmm_1b}
\\
\mathbb{A}_1&=\mathbb{A}'_1/C_1.
\label{eq:ff_fhmm_1c}
\end{align}
Similarly, \emph{subsequent iterations,} i.e., $n=2,\dots,N$, proceed like this
\begin{align}
\mathbb{A}'_n&=\mathbb{G}_n \odot\left( \mathbb{A}_{n-1}\times^{1,2} \Pi^K\cdots \times^{1,2} \Pi^1\right),
\label{eq:ff_fhmm_2a}
\\
C_n&=\left[\mathbb{A}'_n\right],
\label{eq:ff_fhmm_2b}
\\
\mathbb{A}_n&=\mathbb{A}'_n/C_n.
\label{eq:ff_fhmm_2c}
\end{align}
The implementation of this recursive scheme also allows for successive replacement of $\mathbb{G}_n$ with $\mathbb{A}'_n$ and then $\mathbb{A}_n$. This way only one copy of the 3 tensors is required in memory at any given time.

The computational load of each iteration is dominated by the successive tensor-contractions in \cref{eq:ff_fhmm_2a}. Each of those has an asymptotic cost of $O(MM^k)$ per iteration \cite{koldatensor,delathauwer}, which leads to an overall cost of the entire recursive scheme of
\begin{align}
\label{eq:new_complexity}
O\left(N\left(\sum_{k=1}^K MM^k\right)\right)
=
O\left(NM\sum_{k=1}^KM^k\right)
\end{align}
that is a substantial improvement relative to \cref{eq:old_complexity} as it indicates an asymptotic speed-up of $O\left(\prod_{k=1}^KM^k\Big/\sum_{k=1}^KM^k\right)$. In addition, this adapted filtering scheme is also memory efficient, as it operates directly on $\rho^{1:K},\Pi^{1:K}$ avoiding the formation of $\rho,\Pi$ which, for large systems, may cause out-of-memory problems.

\begin{remark}
\label[remark]{rem:HMM_fHMM_comparison}
The computational improvement achieved in the adapted filtering scheme of \cref{eq:ff_fhmm_1a,eq:ff_fhmm_1b,eq:ff_fhmm_1c,eq:ff_fhmm_2a,eq:ff_fhmm_2b,eq:ff_fhmm_2c} relative to \cref{eq:ff_hmm_1a,eq:ff_hmm_1b,eq:ff_hmm_1c,eq:Anprime,eq:Cn,eq:An} is easier to see in the special case where all fHMM subsystems are of the same size $M^1=\cdots=M^K=R$. In this case, naive filtering results in a cost of $O(NR^{2K})$, while adapted filtering results in a cost of only $O(NR^{K+1})$ indicating a speedup of $O(R^{K-1})$.
\end{remark}

\subsubsection{Decoding for fHMM}
\label{sec:viterbi_fHMM}

Similarly to an HMM, the decoding problem of an fHMM requires the computation of the \emph{maximizer} $\tilde s^{1:K}_{1:N}$ of $p(s^{1:K}_{1:N}|w_{1:N},\theta)$ where $\theta=\{\rho^{1:K},\Pi^{1:K},\phi^{1:K},\psi\}$ stands for the parameters of all subsystems. Due to \cref{sec:fHMM_to_HMM_conv}, the decoding of an fHMM is similar to the decoding of an HMM as in \cref{{sec:viterbi_HMM}}. 

In this case, the Viterbi iterations are simplified by the formation of \emph{auxiliary tensors} that are defined by
\begin{align*}
\mathbb{T}_n&=\vect^{-1}\left(\mathcal{T}_n\right)
.
\end{align*}
For the first iteration, this is given directly through the normalized filter
\begin{align}
\label{eq:dec_fHMM_1}
\mathbb{T}_N
=\mathbb{A}_N
\end{align}
and for subsequent iterations, this is given by
\begin{align}
\label{eq:dec_fHMM_2}
\mathbb{T}_n
&=
\mathbb{A}_n\odot 
\mathcal{P}^1\left(\tilde s^1_{n+1}\right)
\odot^1
\cdots
\odot^K
\mathcal{P}^K\left(\tilde s^K_{n+1}\right)
&n&=1,\dots,N-1.
\end{align}
Here, the vectors $\{\{\mathcal{P}^k(\sigma^k)\}_{\sigma^k\in\mathbb{S}^k}\}_{k=1}^K$ are obtained by
\begin{align*}
\mathcal{P}^k\left(\sigma^k\right)
&=
\begin{bmatrix}
\pi^k_{\sigma^k_1\to\sigma^k}
\\
\vdots
\\
\pi^k_{\sigma^k_m\to\sigma^k}
\\
\vdots
\\
\pi^k_{\sigma^k_{M^k}\to\sigma^k}
\end{bmatrix}
,&k&=1,\dots,K.
\end{align*}
These vectors are related to $\mathcal{P}(\sigma)$ according to
\begin{align*}
\mathcal{P}(\sigma)
=
\mathcal{P}^K(\sigma^K)\otimes\cdots\otimes\mathcal{P}^1(\sigma^1).
\end{align*}
Detailed proofs of \cref{eq:dec_fHMM_1,eq:dec_fHMM_2} are provided in the \textsc{supplementary materials.} As with the HMM, the recursion requires $N$ iterations, each consisting of 2 stages, like this
\begin{align*}
\underbrace{
\mathbb{T}_N\to \tilde s^{1:K}_{N}
}_{\text{1\textsuperscript{st} iteration}}
\quad\to\quad
\underbrace{
\mathbb{T}_{N-1}\to \tilde s^{1:K}_{N-1}
}_{\text{2\textsuperscript{nd} iteration}}
\quad\to\quad
\underbrace{
\mathbb{T}_{N-2}\to\tilde s^{1:K}_{N-2}
}_{\text{3\textsuperscript{rd} iteration}}
\quad\to\quad\cdots
.
\end{align*}
The implementation of this recursive scheme allows successive replacement of $\mathbb{T}_{n}$ with $\mathbb{T}_{n-1}$. In this way, only one copy of all auxiliary tensors is required in memory at any given time.

\section{Results}
\label{sec:results}

In this section, we provide characteristic results that compare the performance of the filtering and decoding algorithms through distinct benchmarking tests. To allow for a head-to-head comparison, we use only fHMMs that, due to the equivalence in \cref{sec:fHMM_to_HMM_conv}, can be studied with both vectorized algorithms after conversion to HMM or tensorized algorithms specific to fHMM.

For all tests, we use an fHMM with $K=3$ subsystems, but sizes $M^{1:3},N$ and parameter settings that differ between tests. We carried out our tests using MATLAB (R2026a, update 2) on a standard Apple MacBook Pro (Nov 2024 specs) computer equipped with an Apple M4 Pro chip and 24 GB of memory operated on macOS Tahoe 26.5.1. We conduct all computations in double precision, unless stated otherwise.

\subsection{Filtering algorithms}

First, we compare the numerical accuracy and runtime performance of the vectorized and tensorized filtering schemes.

\subsubsection{Accuracy benchmark test} \label{sec:bench_accuracy}

To assess the numerical accuracy of the two filtering algorithms, we compared the values of the computed marginal likelihoods and filters. Since both algorithms implement the same filtering recursions through different computational approaches, their outputs are expected to agree down to the numerical precision.

To validate this, we conducted all tests with fixed sizes $M^1=2^3,M^2=2^4,M^3=2^5 $ and $N=2^6$ but randomly chosen parameters $\rho^{1:3},\Pi^{1:3}$ and likelihoods $p(w_n|s_n)$. We generated the former via Dirichlet random variates configured to sample uniformly over their respective probability simplexes, and the latter directly via standard uniform random variates. In this way, we ensured that each repetition is performed in a different numerical setting that includes dynamical and observational variations. 

To ensure sufficient sampling of different numerical regimes, we conducted a total of $10^6$ repetitions. In each repetition, we sample new parameters and likelihoods, which we apply to both algorithms and compared the results point-wise via two discrepancy metrics. These are defined by
\begin{align*}
d&=\max_n\left|1-\frac{C^{\rm tens}_n}{C^{\rm vect}_n}\right|
,&
D&=\max_n\max_{\sigma}\left|1-\frac{A^{\rm tens}_n(\sigma)}{A^{\rm vect}_n(\sigma)}\right|
,
\end{align*}
where $C^{\rm vect}_{1:N},\{A^{\rm vect}_{1:N}(\sigma)\}_{\sigma\in\mathbb{S}}$ and $C^{\rm tens}_{1:N},\{A^{\rm tens}_{1:N}(\sigma)\}_{\sigma\in\mathbb{S}}$ denote the constants and filters computed by the forward filtering schemes in \cref{sec:filtering_HMM,sec:filtering_fHMM}, respectively.

As \cref{tab:acc} shows, the resulting maximum discrepancy across all our repetitions was found to be of the order of machine epsilon for both double and single precision computations, indicating that both algorithms produce numerically equivalent marginal likelihoods and filter terms up to floating-point round-off error. These results confirm that our tensorized implementation preserves the numerical accuracy of the standard filtering algorithm.

\begin{table}[htbp]
\caption{Assessment of the numerical accuracy of the vectorized and tensorized filtering algorithms.}
\label{tab:acc}
\centering
\begin{tabular}{lcc}
 & double & single \\
\hline\hline
Maximum $d$ across all repetitions & $1.3323\times10^{-15}$ & $2.4546\times10^{-6}$ \\
Maximum $D$ across all repetitions &  $1.2101\times10^{-14}$ & $7.0333\times10^{-6}$ \\
\hline
Machine $\epsilon$ & $2.2204\times10^{-16}$ & $1.1921\times10^{-7}$ \\
\hline\hline
\end{tabular}
\end{table}

\subsubsection{Runtime benchmark test}
\label{sec:bench_running}

To assess the runtime performance of the two filtering algorithms, we measured their execution time by varying $M^{1:3},N$ one-at-a-time while keeping the others constant at a baseline value. We set such a baseline at $M^1=M^2=M^3=2^4$ and $N=2^6$, which allowed both algorithms to remain safely within the available memory of our computer.

\begin{figure}[H]
    \centering
    \includegraphics[scale = 0.5]{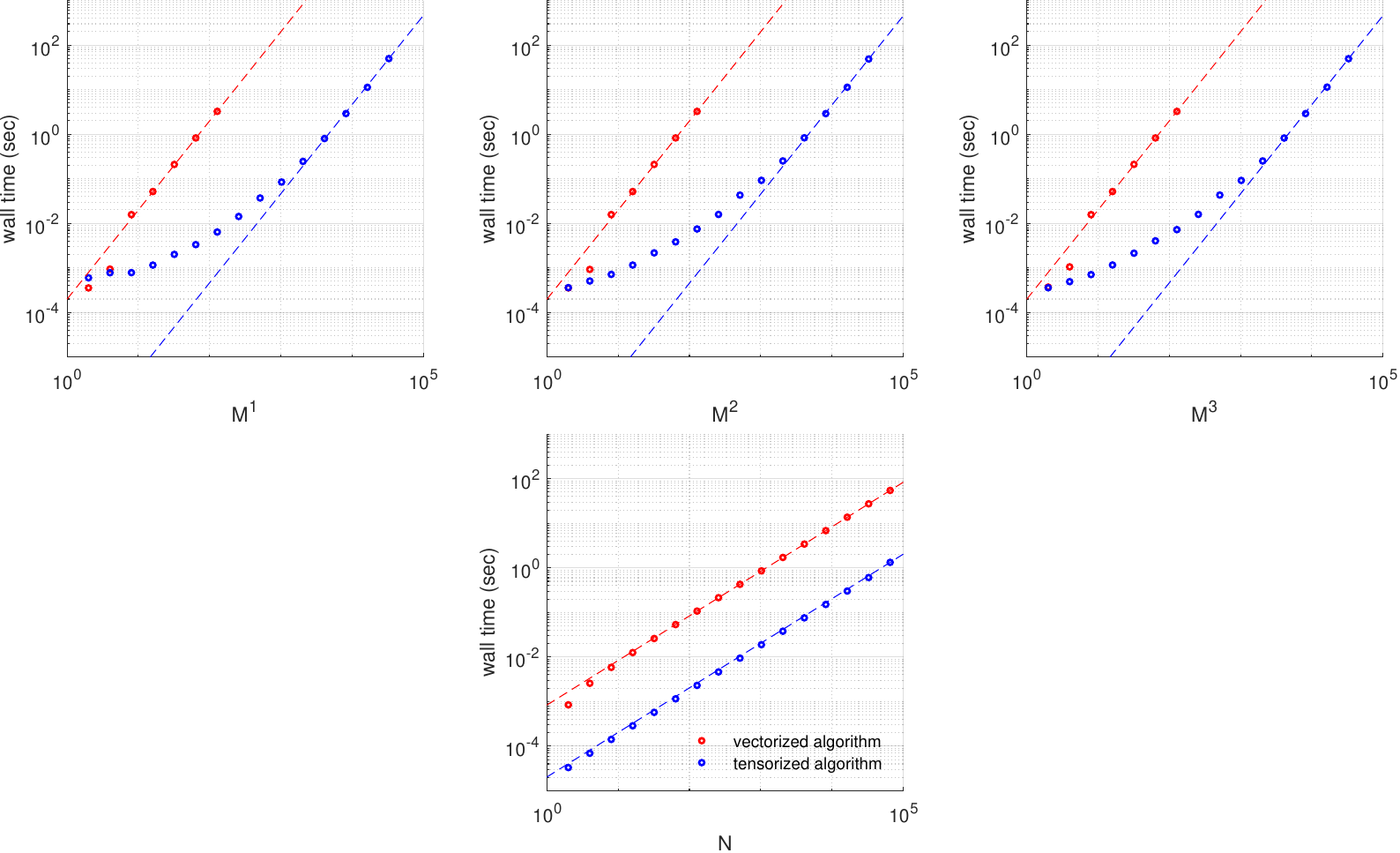}
    \caption{Assessment of the runtime performance of the vectorized and tensorized filtering algorithms. In the panels of this figure only one of the sizes $M^{1:3},N$ varies each time. Dashed lines indicate the theoretically derived asymptotic costs which increase quadratically and linearly for the upper and lower panels, respectively.}
    \label{fig:bench}
\end{figure}

As \cref{fig:bench} shows, the runtimes of both algorithms are in agreement with the asymptotic costs in \cref{eq:old_complexity,eq:new_complexity}. As can be seen, while the runtime naturally increases with all sizes, our tensor-based scheme scales significantly better than the standard vector-based one. This validates the computational gain in the high dimensional setting offered by our novel filtering scheme which asymptotically achieves a speedup of $\approx4500\text{x}$ in the $M^{1:3}$ tests and $\approx45\text{x}$ in the $N$ test.

In addition, this test highlights the \emph{memory efficiency} achieved. Specifically, since our tensorized scheme does not require the construction of $\rho,\Pi$, but instead relies exclusively on the parameters $\rho^{1:3},\Pi^{1:3}$, its memory footprint is drastically lower than that of the vectorized one which proceeds via \cref{eq:kron_rho,eq:kron_Pi}. As a consequence, our novel scheme allowed tests at considerably larger $M^{1:3}$ that may reach as high as $2^{15}$ before the available memory in our computer is exhausted. In contrast, as indicated by the missing points in large $M^{1:3}$ in \cref{fig:bench}, the vectorized algorithm allows only tests up to $2^7$ indicating a much narrowed score of possible applications in size-critical scenarios.

\subsubsection{Additional runtime benchmark test}
\label{sec:bench_equal}

To further assess the runtime performance of the two filtering algorithms, we also measured their execution time in the special case of varying simultaneously $M^{1:3}$. For these tests, we set $M^1=M^2=M^3=R$ for a varying common size $R$ while keeping $N=2^6$ fixed.

\begin{figure}[htbp]
\begin{minipage}[t]{0.48\textwidth}
\centering
\includegraphics[scale=0.25]{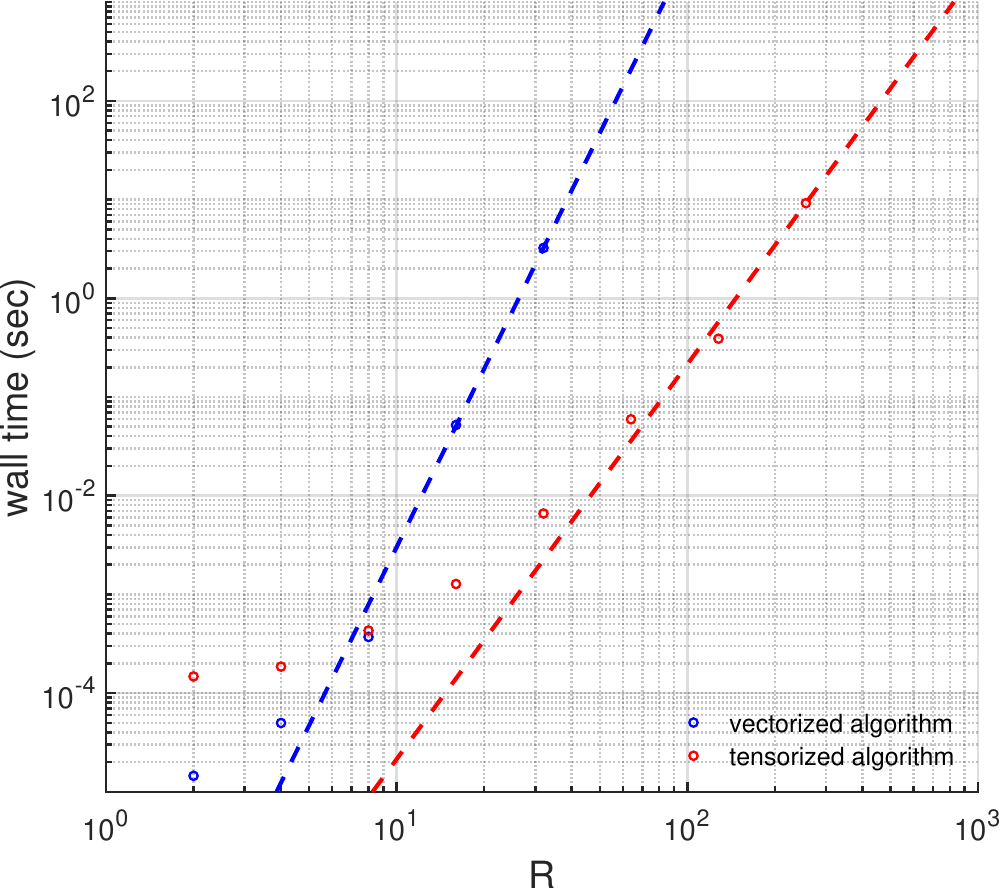}
\caption{Assessment of the runtime performance of the vectorized and tensorized filtering algorithms. In this figure all sizes $M^{1:3}$ are equal to $R$ and vary simultaneously. Dashed lines indicate the theoretically derived asymptotic costs which increase sextically and quartically for the vectorized and tensorized algorithms, respectively.}
\label{fig:bench_equal}
\end{minipage}
\centering
\hfill
\begin{minipage}[t]{0.48\textwidth}
\centering
\includegraphics[scale=0.5]{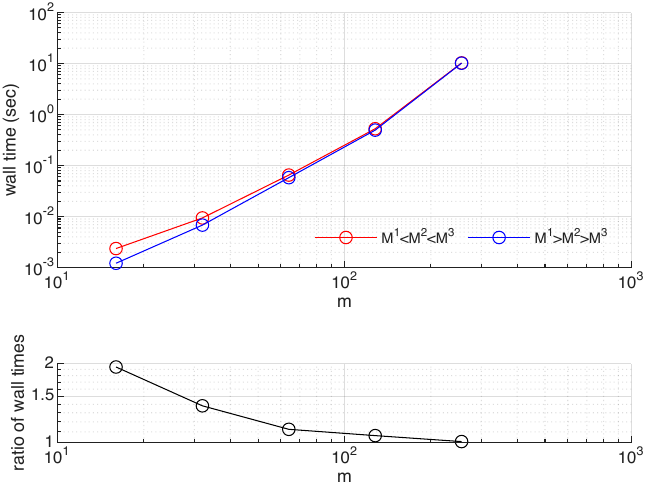}
\caption{Assessment of the runtime performance of the tensorized filtering algorithm only. In this figure, subsystem state-spaces $\mathbb{S}^{1:3}$, of varying configurations, are arranged in an order of ascending or descending size $M^{1:3}$. The filtering runtimes of the resulting fHMM are obtained and compared with each other.}
\label{fig:arrangment}
\end{minipage}

\end{figure}

As \cref{fig:bench_equal} shows, the runtimes of both algorithms, even in this case, remain in agreement with the asymptotic costs $O(R^6)$ and $O(R^4)$ in \cref{rem:HMM_fHMM_comparison}. This verifies once again that the cost of standard vectorized approaches grows very quickly with respect to both the number of subsystems considered and their respective sizes, whereas the cost of the proposed tensorized approaches increases drastically slower.

\subsubsection{Arrangement benchmark test}
\label{sec:bench_order}

In the last filtering benchmark, we only compare the runtime performance of the tensorized algorithm, but in two distinct configurations: one where the fHMM's subsystems are arranged in an order of ascending dimension, i.e., $M^1<M^2<M^3$, and one where the subsystems are arranged in an order of descending dimension, i.e., $M^1>M^2>M^3$.

Specifically, in these tests, we consider a fixed $N=2^6$ and set
\begin{align*}
M^1&=\frac{1}{4}m
,&
M^2&=m
,&
M^3&=4m
\end{align*}
to implement the first configuration, and
\begin{align*}
M^1&=4m
,&
M^2&=m
,&
M^3&=\frac{1}{4}m
\end{align*}
to implement the second, for varying values of the reference size $m$.

As \cref{fig:arrangment} shows, the runtime of both configurations is nearly equal at large sizes; however, at lower or intermediate sizes the second case, i.e.,~$M^1>M^2>M^3$, leads to faster execution times. This indicates that in a practical situation where the fHMM's subsystems are determined by modeling considerations, maximum computational performance can be achieved by placing the largest state-spaces at tensor modes \emph{before} the smallest ones.

\subsection{Decoding algorithms}

To assess the accuracy of the two decoding algorithms, we compared the computed Viterbi maximizers. Since both algorithms rely on numerically equivalent filters (see validation in \cref{sec:bench_accuracy}) and implement similar decoding recursions, it is expected that their output matches.

To validate this, we conducted tests under different numerical regimes similar to those on \cref{sec:bench_accuracy} and, following the application of the appropriate filtering algorithm, we invoked the Viterbi schemes of \cref{sec:viterbi_HMM,sec:viterbi_fHMM} to obtain $\tilde s_{1:N}$ and $\tilde s^{1:K}_{1:N}$, respectively. To compare these sequences, we used the Hamming distance, which measures how different two sequences of equal size are by counting the number of positions where the corresponding elements differ. To interconvert between $\tilde s_{1:N}$ of the HMM and $\tilde s^{1:K}_{1:N}$ of the fHMM representations, we used \cref{eq:revlex_1,eq:revlex_2}. 

From our total $10^6$ repetitions conducted, we observe \emph{no} miss-maching passing state, as identified by the Hamming distances of $0$, indicating that both decoding algorithms produce exactly the same sequences. These results confirm once again that our tensorized implementation preserves the accuracy of the standard algorithm.

\section{Conclusions}
\label{sec:conclusions}

In this study, we presented a unified mathematical formulation that covers both the HMM and fHMM time-series models and developed a novel tensorized version of the forward filtering algorithm that is adapted to fHMMs. Our tensorized scheme allows for scalability across factorial models consisting of a varying number of subsystems, which, until now, has been the main hurtle in the adoption of fHMM for general time-series analysis tasks.

In a series of benchmarking experiments, we compared the standard vectorized with the proposed tensorized algorithm. Both algorithmic schemes yield the same numerical results, but their computational cost, in either execution time or memory, is drastically different, with the adapted out-competing the naive one. 

Future research in time-series modeling and sequential data analysis can build on the methods developed here to enable efficient, stable, and scalable training and inference methods for complex latent-variable state-space models. These may include formulations involving non-linear/non-Gaussian HMMs and their factorial extensions, as well as other models involving similarly structured dynamics. Typically, such models are tackled via Bayesian methods, which are carried out using Markov chain Monte Carlo (MCMC) schemes \cite{liu2001monte,robert2004monte} that exploit conditionally conjugate priors, allowing straightforward Gibbs updates for large blocks of model variables.

However, in practice, domain-driven constraints and specialized modeling scenarios often rule out conjugate priors. For instance, physical considerations may require hard parameter constraints such as positivity, bounded ranges, conservation laws, reversibility and stationarity, or mechanistically motivated likelihoods such as non-Gaussian or non-additive noise or nonlinear observation mappings dictated by instrumentation and measurement equipment. Such modeling features typically break conjugacy and make Gibbs sampling infeasible or inefficient. Consequently, model training, inference, and posterior exploration must rely on more general-purpose MCMC approaches, such as generic Metropolis–Hastings (MH) methods.

However, a key difficulty in all nonconjugate modeling constructions is that both parameter learning and principled model selection hinge on the evaluation of the marginal likelihood, which is expensive and numerically sensitive to compute in high-dimensional models. In this context, the methods developed herein can become an indispensable tool. Specifically, by enabling fast and robust evaluation of the marginal likelihood of factorial models, they provide a practical foundation for: (i) MH-based training procedures, (ii) comparing competing dynamical hypotheses, and (iii) supporting evidence-based decisions about model structure and parameters in realistic data-analysis pipelines.

\bibliographystyle{siamplain}
\bibliography{references}

\end{document}